\documentclass{article} 

\usepackage{amsmath,amsfonts,bm}

\def\eqref#1{equation~\ref{#1}}

\def\1{\bm{1}}

\DeclareMathAlphabet{\mathsfit}{\encodingdefault}{\sfdefault}{m}{sl}
\SetMathAlphabet{\mathsfit}{bold}{\encodingdefault}{\sfdefault}{bx}{n}

\usepackage{ifthen}
\usepackage[normalem]{ulem} 
\usepackage{xcolor}
\newboolean{showedits}
\setboolean{showedits}{true} 
\ifthenelse{\boolean{showedits}}
{
	\newcommand{\del}[1]{\textcolor{red}{\sout{#1}}} 
}{
	\newcommand{\del}[1]{} 
	
}

\newboolean{showcomments}
\setboolean{showcomments}{true} 
\newcommand{\id}[1]{$-$Id: scgPaper.tex 32478 2010-04-29 09:11:32Z oscar $-$}

\ifthenelse{\boolean{showcomments}}
{\newcommand{\nbc}[3]{
		{\colorbox{#3}{\bfseries\sffamily\scriptsize\textcolor{white}{#1}}}
		{\textcolor{#3}{$\blacktriangleright$#2$\blacktriangleleft$}}}
	}
{\newcommand{\nbc}[3]{}
	\renewcommand{\del}[1]{} 
	}

\definecolor{ibcolor}{rgb}{1.0,0.2,.4}
\definecolor{cfcolor}{rgb}{0,0.5,0.9}
\definecolor{oldcolor}{rgb}{0.2,0.2,0.2}
\definecolor{tdcolor}{rgb}{1.0,0,0}
\definecolor{oldcolor}{rgb}{0.5,0.5,0.5}
\definecolor{lycolor}{rgb}{0.3,0.3,0.8}

\usepackage{natbib}
\usepackage{hyperref}
\usepackage{url}
\usepackage{amsmath, amssymb}
\usepackage{graphicx}
\usepackage{algorithm}
\usepackage{algorithmic}
\usepackage[most]{tcolorbox}
\usepackage{subcaption}  
\usepackage{booktabs}    
\usepackage{graphicx}    

\graphicspath{{sec/}} 

\title{Probabilistic Quantum SVM Training on Ising Machine}

\author{Haoqi He \& Yan Xiao \thanks{ Corresponding Author. Address: No. 66, Gongchang Road, Guangming District, Shenzhen, Guangdong 518107, P.R. China  } \\
School of Cyber Science and Technology\\
Shenzhen Campus of Sun Yat-sen University\\
\\
\texttt{hehq23@mail2.sysu.edu.cn \& xiaoyan.hhu@gmail.com}
}

\begin{document}

\maketitle

\begin{abstract}
Quantum computing holds significant potential to accelerate machine learning algorithms, especially in solving optimization problems like those encountered in Support Vector Machine (SVM) training. 
However, current QUBO-based Quantum SVM (QSVM) methods rely solely on binary optimal solutions, limiting their ability to identify fuzzy boundaries in data. Additionally, the limited qubit count in contemporary quantum devices constrains training on larger datasets. %
In this paper, we propose a probabilistic quantum SVM training framework suitable for Coherent Ising Machines (CIMs). 
By formulating the SVM training problem as a QUBO model, we leverage CIMs' energy minimization capabilities and introduce a Boltzmann distribution-based probabilistic approach to better approximate optimal SVM solutions, enhancing robustness. 
To address qubit limitations, we employ batch processing and multi-batch ensemble strategies, enabling small-scale quantum devices to train SVMs on larger datasets and support multi-class classification tasks via a one-vs-one approach.
Our method is validated through simulations and real-machine experiments on binary and multi-class datasets. On the banknote binary classification dataset, our CIM-based QSVM, utilizing an energy-based probabilistic approach, achieved up to 20\% higher accuracy compared to the original QSVM, while training up to $10^4$ times faster than simulated annealing methods. Compared with classical SVM, our approach either matched or reduced training time. On the IRIS three-class dataset, our improved QSVM outperformed existing QSVM models in all key metrics. As quantum technology advances, increased qubit counts are expected to further enhance QSVM performance relative to classical SVM. 

\end{abstract}

\section{Introduction}
\label{sec:intro}

Quantum computing has emerged as a promising technology with significant potential to accelerate machine learning algorithms, particularly in solving NP-hard optimization problems that are challenging for classical computers~\cite{1,2,3,4,5,6,7}. By leveraging quantum parallelism and superposition, quantum computing can efficiently address complex tasks such as convex optimization and RSA cryptanalysis.

SVM are widely used supervised learning algorithms for classification and regression tasks, especially in binary classification~\cite{svmcomplex1,svmcomplex2}. Training an SVM involves solving a convex optimization problem to find the optimal hyperplane that separates different classes in a high-dimensional space, aiming to maximize the margin between classes while minimizing misclassification errors. However, classical SVMs face computational bottlenecks with $O(n^2)$ complexity as data size and feature dimensions grow.

QSVMs have been proposed to alleviate these bottlenecks by utilizing quantum computing devices to solve the convex optimization problem more efficiently. Quantum devices such as CIMs and Quantum Annealers (QAs) offer unique advantages in solving QUBO problems~\cite{qubit1,qubit2,qubit3,qubit4}. By transforming the SVM training problem into a QUBO form, these quantum devices can naturally search for the lowest energy state, efficiently finding optimal or near-optimal solutions.

However, existing QSVM methods face two main challenges:

\textbf{Boundary Ambiguity Problem}: Traditional QUBO models rely solely on binary optimal solutions, making it difficult to handle data with ambiguous boundaries. This limitation leads to insufficient robustness of the model, as slight variations or noise in the data can significantly impact performance~\cite{willsch2020support,bhatia2021performance}.

\textbf{Processing Large-scale Datasets}: The number of qubits in current quantum devices is limited, restricting the ability to train on large datasets. Most gate-model quantum computers are limited to fewer than 100 qubits~\cite{qubit1}, which hinders real-machine experiments and confines research to theoretical and simulation studies. In this paper, we focus on quantum devices such as QA and CIM better for its larger qubits.

To address these challenges, we propose a probabilistic QSVM framework based on CIMs. Our method leverages the energy-based probabilistic sampling rooted in the Boltzmann distribution to enhance the model's ability to handle ambiguous boundaries. By sampling multiple high-probability solutions from the posterior distribution, we construct a probabilistic model that effectively approximates the optimal solution and improves robustness.

Additionally, we introduce batch processing and multi-batch prediction ensemble strategies to cope with the qubit limitations of current quantum hardware. These strategies enable small-scale quantum devices to train SVMs on larger datasets and support multi-class classification tasks using a one-vs-one approach.

The main contributions of this paper are as follows:

\begin{itemize} \item \textbf{We propose a probabilistic QSVM framework suitable for CIMs}: By designing a QSVM model based on an energy framework and probabilistic sampling, we enable efficient training on CIMs and address the limitations of existing QSVMs in handling ambiguous data boundaries.
\item \textbf{We introduce batch processing and multi-batch ensemble strategies}: These methods reduce the number of qubits required for each training session, ensuring that even quantum devices with fewer qubits can complete SVM training on large-scale datasets.

\item \textbf{We validate the effectiveness and performance advantages of CIM-trained SVMs}: Our experiments demonstrate that the proposed QSVM-PROB-CIM outperforms traditional methods, achieving up to a $10^4$ times speedup over simulated annealing in binary classification tasks and improving accuracy by 20\%.

\end{itemize}

\section{Related Work}
\label{sec:related}

In this section, we will show the inspiration and related work of probabilistic based on Boltzmann energy distribution.

\subsection{Quantum Computing and Machine Learning}

Quantum computing has shown great promise in advancing machine learning by providing computational speedups for certain tasks~\cite{rebentrost2014quantum,havlivcek2019supervised}. Quantum devices like CIMs and QAs are particularly effective in solving QUBO problems due to their natural representation of optimization tasks in terms of energy minimization.

\subsection{Traditional QSVM Methods}

\textbf{Quantum Gate Circuit Implementations}: Quantum Support Vector Machines (QSVMs) based on quantum gate models represent data using quantum states and perform matrix operations and inner product calculations through quantum circuits. Rebentrost \emph{et al.}~\cite{rebentrost2014quantum} proposed a quantum version of the SVM algorithm using quantum gates, achieving exponential speedups. Gentinetta \emph{et al.}~\cite{gentinetta2024complexity} proved that variational quantum circuits offer provable exponential speedups in training QSVMs. Havlíček \emph{et al.}~\cite{havlivcek2019supervised} developed quantum kernel algorithms by embedding classical data into quantum state spaces through quantum feature maps, enhancing the expressive power of the model. Jäger and Krems~\cite{jager2023universal} demonstrated that variational quantum classifiers and quantum kernel SVMs based on gate circuits can effectively solve any bounded-error quantum polynomial time problem, highlighting their quantum advantages. Li \emph{et al.}~\cite{li2024quantum} addressed the poor robustness of QSVMs to noisy data by designing new loss functions and iterative optimization strategies implemented on quantum gate circuits.

However, these methods require a large number of qubits and robust quantum error correction, which current quantum gate devices struggle to provide for practical large-scale applications. Most gate-model quantum computers are limited to fewer than 100 qubits~\cite{qubit1}, hindering real-machine experiments and confining research to theoretical and simulation studies.

\textbf{Quantum Annealing and Quantum Approximate Optimization Implementations}: QSVMs based on quantum annealing exploit quantum tunneling effects to solve optimization problems by mapping the SVM objective function to an Ising model or QUBO form. Quantum annealers like the D-Wave systems have been used to tackle quadratic optimization problems. Willsch \emph{et al.}~\cite{willsch2020support} successfully implemented kernel-based QSVMs on the D-Wave 2000Q and validated their effectiveness on small-scale datasets. They suggested that the QA-trained version of SVMs could serve as a practical alternative to classical SVMs. Notably, their work employed actual quantum annealer hardware, confirming the feasibility of training SVMs using QA methods.

However, Bhatia \emph{et al.}~\cite{bhatia2021performance} evaluated the performance of SVMs implemented on D-Wave quantum annealers and pointed out that Willsch's method exhibited high sensitivity to slight variations in the training data. Specifically, the performance could fluctuate significantly in the presence of noisy data or minor perturbations in features. This limitation leads to insufficient robustness of the model, as slight variations or noise in the data can significantly impact performance.

\subsection{QUBO Formulation and Quantum Device Support}

QUBO problems aim to minimize a quadratic objective function with binary variables and are significant in combinatorial optimization. The QUBO form holds special importance in quantum computing because many quantum devices, such as quantum annealers and CIMs, are inherently suited to solve these problems~\cite{quboml}.

Transforming the SVM training problem into a QUBO form allows it to be implemented across various quantum devices. In this mapping, the optimization process of the SVM becomes equivalent to finding the ground state of an Ising model. Specifically, Lagrange multipliers and support vectors are encoded as binary variables, and the SVM's objective function and constraints are converted into quadratic and linear terms in the QUBO formulation. This approach offers several advantages:

\begin{itemize}
    \item \textbf{Broad Device Applicability}: The QUBO form enables SVM training on multiple quantum hardware platforms, including D-Wave quantum annealers and CIMs like those from Qboson~\cite{qboson1,qboson2,qboson3}.

    \item \textbf{Exploitation of Physical Evolution}: Quantum devices naturally search for the lowest energy state through physical processes such as quantum tunneling and coherent superposition, efficiently solving the optimization problem ~\cite{qc1,qc2,qc3}.
\end{itemize}

\subsection{Uncertainty Quantification and Boltzmann Distribution}

In machine learning and deep learning contexts, physical systems that follow the Boltzmann distribution have state probabilities proportional to $e^{-E_i / kT}$~\cite{bm1,bm2,bm3,bm4,bm5}, where $E_i$ is the energy of state $i$, $k$ is the Boltzmann constant, $T$ is the temperature, and $Z$ is the partition function:

\begin{equation}
P(E_i) = \frac{e^{-E_i / kT}}{Z}
\end{equation}

This distribution implies that higher-energy states have lower probabilities, while lower-energy states are more probable. The Boltzmann distribution is crucial in energy minimization and probabilistic modeling, particularly in optimization problems. Classical algorithms such as Restricted Boltzmann Machines, Hopfield networks, simulated annealing, and policy gradient methods utilize this natural probability distribution.

\section{Methodology}
\label{sec:methodology}

In this section, we start from the classic SVM model definition and formulate it as a QUBO problem to achieve the SVM training objective. This energy-based model leverages the Boltzmann distribution to probabilistically utilize multiple sampled solutions, including suboptimal ones, from quantum devices. To overcome the hardware limitations of current quantum devices with a small number of qubits, we employ a batch processing method. By randomly sampling from the original dataset and performing multiple quantum computations, we use batching and statistical sampling techniques to enable small-scale quantum computers to train SVMs on larger datasets. We also extend the binary classification SVM to multi-class tasks.

\subsection{SVM Optimization Problem Definition and QUBO Formulation}
\label{sec:qubo}

The classic SVM training problem seeks the optimal Lagrange multipliers \( \alpha_n \)   by maximizing the following convex objective function: 

\begin{equation}
\max_{\alpha} \left( \sum_{n=1}^N \alpha_n - \frac{1}{2} \sum_{n=1}^N \sum_{m=1}^N \alpha_n \alpha_m y_n y_m K(\mathbf{x}_n, \mathbf{x}_m) \right)
\label{eq:svm_objective}
\end{equation}

Subject to the constraints:

\begin{equation}
\alpha_n \geq 0, \quad \sum_{n=1}^N \alpha_n y_n = 0
\label{eq:svm_constraints}
\end{equation}

Where:

\begin{itemize}
    \item \( \alpha_n \) are the Lagrange multipliers for each sample.
    \item \( y_n, y_m \in \{ \pm 1 \} \) are the labels of the samples.
    \item \( K(\mathbf{x}_n, \mathbf{x}_m) \) is the kernel function between samples \( \mathbf{x}_n \) and \( \mathbf{x}_m \).
    \item \( N \) is the total number of training samples.
\end{itemize}

To adapt this continuous optimization problem for quantum devices like quantum annealers and CIMs, we transform it into a binary optimization problem—a QUBO problem.

We encode each continuous variable \( \alpha_n \) as a sum of binary variables:

\begin{equation}
\alpha_n = \sum_{k=0}^{K-1} B^k a_{Kn + k}
\label{eq:alpha_encoding}
\end{equation}

Where:

\begin{itemize}
    \item \( a_{Kn + k} \in \{0, 1\} \) are binary variables.
    \item \( K \) is the number of bits used for encoding.
    \item \( B \) is the base, typically chosen as \( 2 \) (binary) or \( 10 \).
\end{itemize}

This encoding discretizes each \( \alpha_n \), allowing the optimization problem to be represented entirely with binary variables, conforming to the QUBO formulation.

Substituting the binary representation into the SVM objective function, we obtain the QUBO-formulated objective function:

\begin{equation}
\begin{aligned}
E = & \sum_{n=1}^N \left( \sum_{k=0}^{K-1} B^k a_{Kn+k} \right) - \frac{1}{2} \sum_{n=1}^N \sum_{m=1}^N y_n y_m K(\mathbf{x}_n, \mathbf{x}_m) \\
& \times \left( \sum_{k=0}^{K-1} B^k a_{Kn+k} \right) \left( \sum_{l=0}^{K-1} B^l a_{Km+l} \right)
\end{aligned}
\label{eq:qubo_objective}
\end{equation}

In this objective function, each \( \alpha_n \) is replaced by its binary representation, and the function is entirely expressed in terms of binary variables \( a_{Kn+k} \), conforming to the QUBO form. 

To incorporate the SVM constraint \( \sum_{n=1}^N \alpha_n y_n = 0 \), we introduce a penalty term to the objective function:

\begin{equation}
E_{\text{total}} = E + \xi \left( \sum_{n=1}^N \left( \sum_{k=0}^{K-1} B^k a_{Kn+k} \right) y_n \right)^2
\label{eq:total_energy}
\end{equation}

Where \( \xi \) is a positive penalty coefficient that controls the strictness of the constraint. Due to the also adopted QUBO formulation, this form is in fact similar to our object of comparison D-Wave QSVM ~\cite{willsch2020support}.

By adding this quadratic penalty term, the constraint is embedded into the objective function, guiding the quantum optimization process to find solutions that satisfy the constraint, making the final solution closer to that of the original SVM optimization problem.






\subsection{Boltzmann Distribution and Probabilistic Quantum SVM}
\label{sec:boltzmann}

In classical SVM training, the optimal solution corresponds to continuous optimal values of the Lagrange multipliers \( \alpha_n \). However, in Quantum SVMs, since quantum devices solve the QUBO problem yielding discrete solutions, directly using these discrete solutions may lead to results deviating from the SVM's optimal solution (as observed in our experiments and previous studies). To address this, we introduce a probabilistic method based on the Boltzmann distribution to convert the discrete QUBO solutions into probabilistic weights, thereby better approximating the true optimal solution of the SVM.

Specifically, we assign a probability to each solution based on the Boltzmann distribution from thermodynamic statistics, ensuring that solutions with lower energy (closer to the optimal solution) have higher probabilities.

During the QUBO problem-solving process, the set of solutions \( \{ x_i \} \) corresponds to different energies \( E_i \). We define the partition function \( Z \) to normalize the probabilities:

\begin{equation}
Z = \sum_{i=1}^{M} e^{-E_i / kT}
\label{eq:partition_function}
\end{equation}

Where:

\begin{itemize}
    \item \( M \) is the total number of solutions.
    \item \( E_i \) is the energy of the \( i \)-th solution.
    \item \( k \) is the Boltzmann constant.
    \item \( T \) is the temperature parameter.
\end{itemize}

The probability weight \( P(x_i) \) for each solution \( x_i \) is calculated as:

\begin{equation}
P(x_i) = \frac{e^{-E_i / kT}}{Z}
\label{eq:boltzmann_probability}
\end{equation}

This probability distribution ensures that solutions with lower energy have higher probabilities, favoring the capture of solutions close to the optimal one.

To better simulate the true optimal solution of the SVM, we convert the discrete QUBO solutions into probabilistic continuous values for the Lagrange multipliers \( \alpha_n \).

Assuming that we have obtained a set of solutions represented as binary vectors \( X = [x_1, x_2, \ldots, x_M] \), where each solution \( x_i \) corresponds to a possible combination of support vector multipliers.

We compute the probability-weighted Lagrange multipliers \( \alpha_n \):

\begin{equation}
\alpha_n = \sum_{i=1}^{M} x_{i,n} \cdot P(x_i)
\label{eq:probabilistic_alpha}
\end{equation}

Where:

\begin{itemize}
    \item \( x_{i,n} \) is the value of the binary variable corresponding to \( \alpha_n \) in solution \( x_i \).
    \item \( P(x_i) \) is the probability weight of solution \( x_i \).
\end{itemize}

By summing the discrete solutions \( x_i \) weighted by their probabilities, we obtain continuous values for \( \alpha_n \) that are closer to the form of the SVM's optimal solution.

To further approximate the optimal decision boundary of the SVM, we compute the bias term \( b \) based on the probabilistically obtained \( \alpha \) values:

\begin{equation}
b = \frac{\sum_{n=1}^{N} \alpha_n (C - \alpha_n) \left( y_n - \sum_{m=1}^{N} \alpha_m y_m K(\mathbf{x}_m, \mathbf{x}_n) \right)}{\sum_{n=1}^{N} \alpha_n (C - \alpha_n)}
\label{eq:probabilistic_bias}
\end{equation}

Where:

\begin{itemize}
    \item \( y_n \) is the label of the \( n \)-th sample.
    \item \( C \) is the SVM penalty parameter.
    \item \( K(\mathbf{x}_m, \mathbf{x}_n) \) is the kernel function defining the similarity between samples. We usually use a Gaussian kernel.
\end{itemize}

Through this method, we obtain \( \alpha \) and \( b \) that are closer to the true optimal solution of the SVM, resulting in better classification performance on the data.

By employing this probabilistic approach based on the Boltzmann distribution, we effectively transform the discrete solutions from quantum computing into probabilistic solutions that align with the optimal characteristics of SVM solutions, providing a new avenue for efficient machine learning implementations on quantum computing devices.

\subsection{Training on Large Datasets with Small Quantum Devices}
\label{sec:batch}

Focusing on binary classification tasks, for datasets smaller than the qubit limit of the quantum device, we can process them in one go to maximize quantum advantages. However, to overcome the hardware constraints of current small-scale noisy quantum devices with limited qubits, we adopt batch processing and multi-batch prediction ensemble methods to improve the current QSVM. Similar ideas can be found in numerous fields related to machine learning and quantum computing ~\cite{small1,small2,small3,small4}.

Firstly, the batch processing mechanism divides the original dataset into small batches, each with a data size less than the qubit limit. This strategy reduces the number of samples and qubits required for each computation.

For each pair of classes in the dataset, we divide the data into multiple batches, with the batch size controlled by a parameter \( B \). The original dataset \( \mathcal{D} = \{ (\mathbf{x}_i, y_i) \}_{i=1}^N \) is partitioned into \( n_b = \lceil N / B \rceil \) small batches, each containing \( B \) samples (i.e., \( B \leq \text{qubits limitation} \)):

\begin{equation}
\mathcal{D} = \bigcup_{j=1}^{n_b} \mathcal{D}_j, \quad \mathcal{D}_j = \{ (\mathbf{x}_{i_j}, y_{i_j}) \}_{i_j=1}^B
\label{eq:batch_partition}
\end{equation}

We train one batch at a time. After training each batch, we store the corresponding Lagrange multipliers \( \alpha \) and bias term \( b \) for the binary classification model, which are used for subsequent predictions, thereby gradually completing the training of the entire dataset.

For each batch of data, we use the quantum SVM model to obtain the Lagrange multipliers \( \alpha \) and bias term \( b \). Specifically, for the \( j \)-th batch \( \mathcal{D}_j \), the obtained Lagrange multipliers are \( \alpha_j \) and the bias term is \( b_j \):

\begin{equation}
\alpha_{j,n} = \sum_{k=0}^{K-1} B^k a_{Kn + k}, \quad b_j = \text{function}(\alpha_j, K(\mathbf{x}_m, \mathbf{x}_n))
\label{eq:batch_alpha_bias}
\end{equation}

By dividing the dataset into multiple small batches, each training session requires only a small number of qubits, trading off more computation rounds for reduced qubit requirements.

Next, we employ a multi-batch prediction ensemble approach. In the prediction phase, we aggregate the predictions from multiple batch models for each class pair to enhance classification stability.

For each binary classification model between classes \( (i, j) \), we retrieve all batch parameters \( (\alpha_j, b_j) \) and perform predictions batch by batch, obtaining the prediction results for each batch:

\begin{equation}
\hat{y}_j = \text{sign}\left( \sum_{m=1}^B \alpha_j^{(m)} y_m K(\mathbf{x}_m, \mathbf{x}_{\text{test}}) + b_j \right)
\label{eq:batch_prediction}
\end{equation}

For each binary classification model, we collect the predictions from all batches and compute the average:

\begin{equation}
\bar{y} = \frac{1}{n_b} \sum_{j=1}^{n_b} \hat{y}_j
\label{eq:ensemble_prediction}
\end{equation}

We use the averaged prediction result as the final output of the binary classification model. If \( \bar{y} > 0 \), the sample is predicted as class \( i \); otherwise, it is predicted as class \( j \).

By averaging the prediction results from different batches, we mitigate the randomness of single-batch predictions, enhancing the robustness and accuracy of the model.

\subsection{One-vs-One Multi-class Probabilistic Quantum SVM}
\label{sec:multiclass}

The one-vs-one strategy decomposes multi-class problems into multiple binary classification problems, extending the original binary QSVM to multi-class tasks. The binary QSVM employs the batch processing and multi-batch prediction ensemble methods described in Section~\ref{sec:batch}, reducing the data size and thereby lowering the required number of qubits.

For every pair of classes \( (i, j) \) in the dataset, we select only the data belonging to these two classes and convert the class labels to binary labels (\(+1\) and \(-1\)). We then use a voting mechanism where the prediction results of all binary classifiers serve as votes to assign the sample to the class with the most votes.

For each class pair \( (i, j) \), we construct a sub-dataset:

\begin{equation}
\mathcal{D}_{ij} = \{ (\mathbf{x}_n, y_n) \mid y_n = i \text{ or } y_n = j \}
\label{eq:pairwise_dataset}
\end{equation}

We convert the labels \( y_n \) to binary labels: \( y_n = +1 \) for class \( i \), and \( y_n = -1 \) for class \( j \).

For each class pair \( (i, j) \), we use the trained binary classification model to predict the test sample \( \mathbf{x}_{\text{test}} \):

\begin{equation}
\hat{y}_{ij} = \text{sign}\left( \sum_{n=1}^B \alpha_{ij}^{(n)} y_n K(\mathbf{x}_n, \mathbf{x}_{\text{test}}) + b_{ij} \right)
\label{eq:pairwise_prediction}
\end{equation}

We collect the votes for each class from all binary classifiers. If \( \hat{y}_{ij} > 0 \), class \( i \) receives one vote; otherwise, class \( j \) receives one vote.

The test sample is assigned to the class with the highest number of votes:

\begin{equation}
\hat{y} = \arg\max_{k} \text{votes}(k)
\label{eq:voting}
\end{equation}

This voting mechanism ensures that the prediction results of all binary classifiers jointly determine the final classification result in multi-class tasks, enhancing the decision reliability of the model.


The batch processing, multi-batch prediction ensemble, and one-vs-one multi-class strategies proposed in this section effectively overcome the limitations of current quantum devices in terms of qubit count. They enable quantum SVMs to train and predict efficiently on multi-class tasks and larger datasets, providing new solutions for the practical application of quantum machine learning.

\begin{algorithm}[ht]
\caption{Probabilistic Quantum SVM with Batch Processing and Multi-Class Extension}
\label{alg:quantum_svm}
\begin{algorithmic}[1]
\STATE \textbf{Input:} Dataset $\mathcal{D} = \{ (\mathbf{x}_i, y_i) \}_{i=1}^N$, kernel $K$, temperature $T$, batch size $B$, classes $C$
\STATE \textbf{Output:} Trained model parameters $\alpha$, $b$

\STATE \textbf{1. SVM Optimization as QUBO} 
\STATE Define the SVM objective function and transform $\alpha_n$ to binary variables, reformulating the objective as a QUBO problem.

\STATE \textbf{2. Solve QUBO on Quantum Device} 
\STATE Use the quantum device to obtain solutions $\{ x_i \}$ with corresponding energies $\{ E_i \}$.

\STATE \textbf{3. Boltzmann-based Probabilistic Weighting} 
\STATE Calculate the partition function $Z$ and assign probability weights $P(x_i)$ based on energy levels. Compute weighted Lagrange multipliers $\alpha_n$ and bias term $b$.

\STATE \textbf{4. Batch Processing for Large Datasets} 
\STATE Partition dataset $\mathcal{D}$ into batches $\{ \mathcal{D}_j \}$ of size $B$. 
\FOR{each batch $\mathcal{D}_j$}
    \STATE Train the quantum SVM on batch $\mathcal{D}_j$ to obtain parameters $\alpha_j$, $b_j$.
    \STATE Store $\alpha_j$, $b_j$ for ensemble prediction.
\ENDFOR

\STATE \textbf{5. Multi-Batch Ensemble Prediction} 
\FOR{each test sample $\mathbf{x}_{\text{test}}$}
    \STATE Aggregate predictions from all batches and compute the final averaged prediction $\bar{y}$.
\ENDFOR

\STATE \textbf{6. One-vs-One Multi-Class Classification} 
\FOR{each class pair $(i, j)$}
    \STATE Construct binary-labeled subset $\mathcal{D}_{ij}$.
    \STATE Train binary classifier on $\mathcal{D}_{ij}$.
    \FOR{each test sample $\mathbf{x}_{\text{test}}$}
        \STATE Predict $\hat{y}_{ij}$.
    \ENDFOR
    \STATE Use voting to assign the final class.
\ENDFOR

\STATE \textbf{Return:} $\alpha$, $b$ and trained multi-class model.
\end{algorithmic}
\end{algorithm}

The pseudocode in Algorithm \ref{alg:quantum_svm} presents our probabilistic quantum SVM framework, focusing on an innovative Boltzmann distribution-based probabilistic approach to optimize quantum SVM solutions. After defining the SVM optimization as a QUBO problem in lines 1.1–1.4, we transition from a conventional binary solution approach to a probabilistic solution by incorporating Boltzmann weighting. This shift is crucial: while conventional quantum SVMs often rely on single, binary optimal solutions that can lack robustness, our method takes advantage of multiple quantum-generated solutions with varying energy levels.

Lines 2–3 detail this probabilistic enhancement using the Boltzmann distribution. In line 2.1, we use the quantum device to produce a set of solutions $\{x_i\}$, each associated with an energy $E_i$. These solutions reflect different potential configurations of the Lagrange multipliers, with lower energies indicating configurations closer to the optimal solution. To leverage these multiple solutions, line 3.1 calculates the partition function $Z$, normalizing the energy-based probabilities. Lines 3.2–3.3 assign a probability weight $P(x_i)$ to each solution, increasing the influence of low-energy solutions by using the Boltzmann distribution as a filter. This probabilistic weighting ensures that configurations closer to the optimal energy contribute more significantly to the final SVM solution, effectively blending suboptimal solutions to create a continuous, robust approximation of the optimal SVM decision boundary.

The probabilistic Lagrange multipliers $\alpha_n$, calculated in line 3.4 as the weighted sum over all solutions, allow the quantum SVM to model complex, noisy, or ambiguous data points more effectively. By not solely relying on a single optimal configuration, our framework mimics the natural probabilistic behavior of thermal systems, capturing a range of near-optimal solutions that better approximate the SVM objective. The bias term $b$ (line 3.5), derived from these probabilistic $\alpha$ values, completes the decision boundary approximation, resulting in a classifier that more closely aligns with the true continuous solution of classical SVMs. This probabilistic approach, enabled by Boltzmann distribution weighting, stands out as a unique adaptation for quantum SVMs, overcoming the rigidity of binary-only solutions and providing robustness to the model’s predictions.

Batch processing (lines 4–4.3) further extends the method’s applicability by dividing the dataset $\mathcal{D}$ into manageable batches $\{ \mathcal{D}_j \}$, allowing for training with limited qubits. The ensemble prediction in lines 5–5.3 stabilizes predictions by averaging across batches, and lines 6–6.3 employ a one-vs-one strategy for multi-class classification. Through this process, the Boltzmann-based probabilistic sampling, batch processing, and multi-class extensions combine to deliver an effective quantum SVM approach with heightened robustness and flexibility.

\section{Experiments}
\label{sec:experiments}

In this section, we evaluate our probabilistic quantum SVM (QSVM-PROB-CIM) and compare it with previous quantum SVM implementations and classical SVMs. We conducted experiments on both binary and multi-class datasets using a simulated CIM and performed additional training on a real CIM provided by Beijing Boson Quantum Technology Co., Ltd. The comparison includes SVM implementations on D-Wave quantum annealers and classical SVMs using the \texttt{scikit-learn} library ~\cite{scikit1,sklearn2}. Methods marked with ``PROB'' utilize our proposed probabilistic approach, and those labeled ``Real Machine'' indicate training performed on actual quantum hardware such as Qboson CIM 550W which has 550 qubits.

In the tables presented, ``N/A'' denotes cases where the number of qubits is not applicable, typically for classical SVMs.

\subsection{Experiment 1: Binary Classification on Banknote Dataset}

We tested binary classification tasks using two datasets, \textit{banknote\_1} and \textit{banknote\_2}, randomly sampled from the Banknote Authentication dataset. Each dataset contains 250 training points and 100 validation points. The common parameters C and gamma of SVM are set to 3 and 16, the common B and K of QSVM are set to 2, and the penalty is set to 0.001.

\begin{figure*}[htbp]
    \centering
    \begin{subfigure}{0.48\linewidth}
        \centering
        \includegraphics[width=\linewidth]{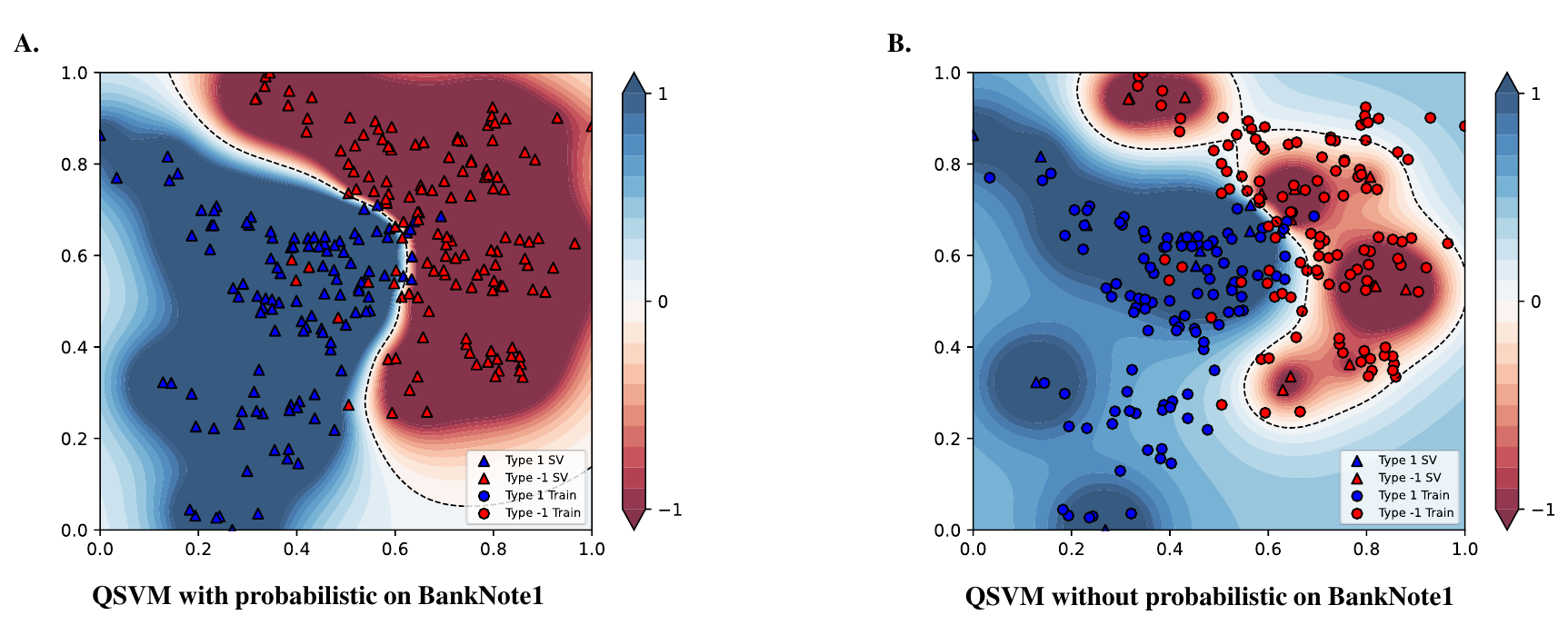}
        \caption{The visualize on \textit{banknote\_1} dataset}
        \label{fig:1}
    \end{subfigure}
    \hfill
    \begin{subfigure}{0.48\linewidth}
        \centering
        \includegraphics[width=\linewidth]{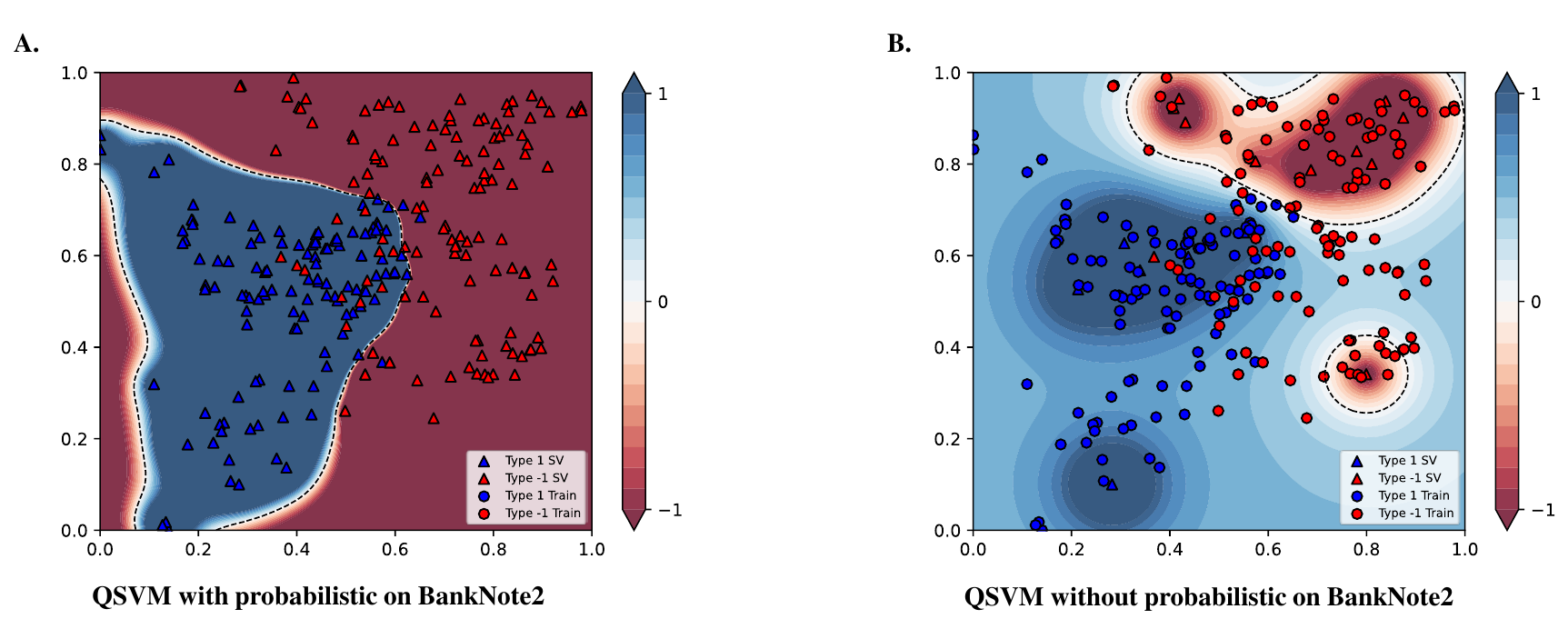}
        \caption{The visualize on \textit{banknote\_2} dataset}
        \label{fig:2}
    \end{subfigure}
    \caption{Visualizing the effect of probabilistic on boundaries}
    \label{fig:combined}
\end{figure*}

Figure \ref{fig:combined} illustrates the impact of the probabilistic approach on decision boundaries across the \textit{banknote\_1} and \textit{banknote\_2} datasets. This composite figure comprises two subfigures, (a) and (b), where subfigure A in each panel represents the decision boundary generated with the probabilistic method applied, while subfigure B shows the boundary without it. Through visual analysis, it becomes evident that the probabilistic method induces a more pronounced adjustment to the decision boundary, particularly in regions with dense data points. The subtle shifts in the boundary reveal the algorithm's capacity to leverage near-optimal solutions. This probabilistic strategy enhances the model’s robustness and adaptability in handling complex boundaries. Furthermore, the probabilistic approach smooths the boundary, markedly improving classification performance for edge data points. By introducing flexibility at the boundary, this method allows the model to manage outliers with greater agility, avoiding rigid splits and thereby enhancing the overall accuracy of the model.

\begin{table}[h]
\centering
\caption{Comparison of SVMs using Energy-Based Probabilistic Methods on \textit{banknote\_1}}
\label{tab:banknote1}
\resizebox{\columnwidth}{!}{
\begin{tabular}{lcccccc}
\toprule
\textbf{Metrics} & \textbf{CSVM} & \textbf{QSVM-} & \textbf{SVM-SA} & \textbf{QSVM-} & \textbf{QSVM-} & \textbf{QSVM-PROB-} \\
 &  & \textbf{Qiskit} &  & \textbf{D-Wave} & \textbf{PROB-CIM} & \textbf{CIM Real} \\
\midrule
Accuracy          & 0.94 & 0.63   & 0.96 & 0.76 & 0.96 & \textbf{0.96}\\
Precision         & 0.94 & 0.5778 & 0.9333 & 1.0 & 0.9778 & \textbf{0.9778}\\
Recall            & 0.94 & 0.5909 & 0.9767 & 1.0 & 0.9362 & \textbf{0.9362}\\
F1 Score          & 0.94 & 0.5843 & 0.9545 & 0.77 & 0.9565 & \textbf{0.9565}\\
Training Time     & 2.99 ms & 1150.32 s & 11.11 s & 3.71 ms & 1.23 s & \textbf{2.71 ms}\\
Quantum Resources & N/A  & 4       & N/A    & 2000  & 550   & \textbf{550}\\
\bottomrule
\end{tabular}
}
\end{table}

\begin{table}[h]
\centering
\caption{Comparison of SVMs using Energy-Based Probabilistic Methods on \textit{banknote\_2}}
\label{tab:banknote2}
\resizebox{\columnwidth}{!}{
\begin{tabular}{lcccccc}
\toprule
\textbf{Metrics} & \textbf{CSVM} & \textbf{QSVM-} & \textbf{SVM-SA} & \textbf{QSVM-} & \textbf{QSVM-} & \textbf{QSVM-PROB-} \\
 &  & \textbf{Qiskit} &  & \textbf{D-Wave} & \textbf{PROB-CIM} & \textbf{CIM Real} \\
\midrule
Accuracy          & 0.93 & 0.61   & 0.98 & 0.80 & 0.95 & \textbf{0.95}\\
Precision         & 0.93 & 0.5750 & 0.9722 & 1.0 & 0.9444 & \textbf{0.9444}\\
Recall            & 0.93 & 0.5111 & 0.9722 & 0.90 & 0.9189 & \textbf{0.9189}\\
F1 Score          & 0.93 & 0.5412 & 0.9722 & 0.75 & 0.9315 & \textbf{0.9315}\\
Training Time     & 3.43 ms & 1149.63 s & 11.98 s & 3.89 ms & 1.29 s & \textbf{2.49 ms}\\
Quantum Resources & N/A  & 4       & N/A    & 2000  & 550   & 550 \\
\bottomrule
\end{tabular}
}
\end{table}

From Tables~\ref{tab:banknote1} and~\ref{tab:banknote2}, we observe the following:

\begin{itemize}
    \item \textbf{Classical SVM (CSVM)} serves as a baseline, achieving high performance across all metrics with short training times due to optimized library functions.
    \item \textbf{QSVM-Qiskit}, representing gate-model quantum kernel SVMs, utilizes a small number of superconducting qubits. However, it shows lower accuracy and longer simulation times, highlighting performance limitations.
    \item \textbf{SVM-SA}, trained using simulated annealing, demonstrates good performance but has significantly longer training times compared to CSVM.
    \item \textbf{QSVM-PROB-CIM} outperforms the basic QSVM-CIM by incorporating the energy-based probabilistic method. In \textit{banknote\_1}, accuracy improves by 20\%, and in \textit{banknote\_2}, there is a notable enhancement across all metrics.
    \item \textbf{QSVM-PROB-CIM Real Machine} shows that real CIM hardware can achieve performance comparable to simulations, with even shorter training times, showcasing the practical potential of CIMs.
    \item \textbf{QSVM-D-Wave} and \textbf{QSVM-PROB D-Wave} indicate that while D-Wave quantum annealers can perform the tasks, their performance is not as competitive as CIM-based methods, and training times are longer.
\end{itemize}

The training speeds on quantum hardware like D-Wave and CIM are close to or slightly faster than the optimized CSVM, demonstrating the potential of quantum devices in accelerating machine learning model training. As quantum computing technology advances and the number of qubits increases, QSVMs are expected to surpass CSVMs on large datasets due to their quadratic time complexity advantage.

\subsection{Experiment 2: Multi-class Classification on IRIS Dataset}

In this experiment, we evaluated the performance on the IRIS dataset, a well-known three-class dataset in \texttt{scikit-learn} library. The data volume is within the capacity of quantum devices with around 100 qubits. The experiment takes 40\% of the training set with all parameters set to 1, with 42 random seeds.

\begin{table}[h]
\centering
\caption{Comparison of Energy-Based Probabilistic Methods in Multi-class SVM on IRIS Dataset}
\label{tab:iris}
\resizebox{\columnwidth}{!}{
\begin{tabular}{lccccc}
\toprule
\textbf{Metrics} & \textbf{CSVM} & \textbf{SVM-SA} & \textbf{QSVM-} & \textbf{QSVM-PROB-} & \textbf{QSVM-PROB-} \\
 &  &  & \textbf{CIM} & \textbf{CIM} & \textbf{CIM Real} \\
\midrule
Accuracy          & 0.9833 & 0.9833 & 0.30  & 0.9833 & \textbf{0.9833}\\
Precision         & 0.9833 & 0.9833 & 0.10  & 0.9833 & \textbf{0.9833}\\
Recall            & 0.9800 & 0.9800 & 0.3333 & 0.9800 & \textbf{0.9800}\\
F1 Score          & 0.9800 & 0.9800 & 0.1533 & 0.9800 & \textbf{0.9800}\\
Training Time     & 6.1 ms & 30.51 s & 0.244 s & 0.234 s & \textbf{5.933 ms}\\
Quantum Resources & N/A    & N/A     & 100    & 100     & \textbf{100}\\
\bottomrule
\end{tabular}
}
\end{table}

\noindent\textbf{Note}: Test size is 40\% of the dataset, with a random state of 42. Three separate trainings were conducted, each using a one-vs-one approach.

From Tables~\ref{tab:iris}, we observe the following:
\begin{itemize}
    \item Both \textbf{CSVM} and \textbf{SVM-SA} perform exceptionally well on the multi-class task, achieving nearly perfect scores across all metrics.
    \item The basic \textbf{QSVM-CIM} model achieves an accuracy of only 0.30, indicating its insufficiency in handling multi-class tasks without enhancements.
    \item \textbf{QSVM-PROB-CIM} significantly improves performance by employing the probabilistic method, matching the high accuracy and F1 scores of CSVM.
    \item The \textbf{QSVM-PROB-CIM Real Machine} achieves comparable performance to simulations with even shorter training times, highlighting the acceleration capabilities of CIM hardware in multi-class tasks.
\end{itemize}

These results demonstrate that the probabilistic method effectively enhances the QSVM's ability to handle multi-class classification, bringing its performance on par with classical models.

\subsection{Experiment 3: Ablation Study}

To further validate the effectiveness of our probabilistic method, we conducted an ablation study comparing QSVMs with and without the probabilistic enhancement on both CIM and D-Wave platforms.
\begin{table}[h]
\centering
\caption{Ablation Study Results on \textit{banknote\_1}}
\label{tab:ablation1}
\resizebox{\columnwidth}{!}{
\begin{tabular}{lcccccc}
\toprule
\textbf{Metrics} & \textbf{QSVM-} & \textbf{QSVM-\textbf{PROB}-} & \textbf{QSVM-} & \textbf{QSVM-\textbf{PROB}-} & \textbf{QSVM-} & \textbf{QSVM-\textbf{PROB}} \\
 & \textbf{CIM} & \textbf{CIM} & \textbf{CIM Real} & \textbf{CIM Real} & \textbf{D-Wave} & \textbf{D-Wave} \\
\midrule
Accuracy      & 0.76 & \textbf{0.96} & 0.96 & \textbf{0.96} & 0.76 & \textbf{0.96} \\
Precision     & 1.0  & \textbf{0.9778} & 0.9778 & \textbf{0.9778} & 1.0 & \textbf{0.9333} \\
Recall        & 0.6522 & \textbf{0.9362} & 0.9362 & \textbf{0.9362} & 1.0 & \textbf{0.9767} \\
F1 Score      & 0.7895 & \textbf{0.9565} & 0.9565 & \textbf{0.9565} & 0.77 & \textbf{0.9545} \\
\bottomrule
\end{tabular}
}
\end{table}

\begin{table}[h]
\centering
\caption{Ablation Study Results on \textit{banknote\_2}}
\label{tab:ablation2}
\resizebox{\columnwidth}{!}{
\begin{tabular}{lcccccc}
\toprule
\textbf{Metrics} & \textbf{QSVM-} & \textbf{QSVM-\textbf{PROB}-} & \textbf{QSVM-} & \textbf{QSVM-\textbf{PROB}-} & \textbf{QSVM-} & \textbf{QSVM-\textbf{PROB}} \\
 & \textbf{CIM} & \textbf{CIM} & \textbf{CIM Real} & \textbf{CIM Real} & \textbf{D-Wave} & \textbf{D-Wave} \\
\midrule
Accuracy      & 0.94 & \textbf{0.95} & 0.87 & \textbf{0.95} & 0.80 & \textbf{0.98} \\
Precision     & 0.9167 & \textbf{0.9444} & 0.7222 & \textbf{0.9444} & 1.0 & \textbf{0.9722} \\
Recall        & 0.9167 & \textbf{0.9189} & 0.8965 & \textbf{0.9189} & 0.90 & \textbf{0.9722} \\
F1 Score      & 0.9167 & \textbf{0.9315} & 0.8000 & \textbf{0.9315} & 0.75 & \textbf{0.9722} \\
\bottomrule
\end{tabular}
}
\end{table}

From Tables~\ref{tab:ablation1} and ~\ref{tab:ablation2}, we observe the following:
\textbf{QSVM-CIM vs. QSVM-PROB-CIM}: The energy-based probabilistic method significantly enhances performance across all metrics. For instance, in Table~\ref{tab:ablation1}, accuracy increases from 0.76 to 0.96, and the F1 score rises from 0.7895 to 0.9565, demonstrating the substantial impact of the probabilistic method on model performance.

\textbf{Real Machine vs. Simulation}: QSVM-PROB-CIM achieves comparable performance to simulation on real CIM hardware, with a 10,000-fold improvement in training speed, and is also comparable to classical implementations, indicating the practical potential and efficiency of CIM. The quantum annealer of D-Wave has the same comparative conclusion.

Our experiments demonstrate the successful training of QSVMs using CIM, highlighting the potential practicality of quantum devices like CIM in machine learning. The real-machine experiments further showcase the acceleration capabilities of CIM in training machine learning models. Moreover, our probabilistic method enhances model performance by correcting the QUBO solution set using suboptimal solutions, outperforming traditional methods that directly use the optimal solution. These improvements indicate that small quantum devices can effectively train multi-class models and handle large datasets.


\section{Conclusion}
\label{sec:Conclusion}

The experimental results indicate that our QSVM based on CIM and the probabilistic method exhibit excellent performance. As the number of qubits increases, the QSVM using CIM is expected to demonstrate advantages on larger datasets compared with classic SVM's quadratic-level complexity. In future developments, as quantum computing technology progresses, the application of them in machine learning is expected to expand further, potentially surpassing classical SVMs in handling complex tasks. Our work lays a foundation for practical quantum machine learning implementations and highlights the importance of leveraging probabilistic methods to enhance quantum algorithms. While our experiments show promising results, current quantum devices still face limitations in qubit counts and error rates. Future work will focus on scaling our approach to larger datasets and exploring error mitigation techniques to further improve performance. Additionally, extending the probabilistic method to other quantum machine learning models could open new avenues for research.

\section*{Acknowledgments}
This research was supported by the Management Science \& Engineering-Bose Quantum Beijing Inc. Foundation 2024.

\bibliography{iclr2025_conference}
\bibliographystyle{iclr2025_conference}

\appendix

\end{document}